\documentclass[letterpaper, 10 pt, conference]{ieeeconf}  % Comment this line out if you need a4paper

\IEEEoverridecommandlockouts                              % This command is only needed if
\makeatletter
\let\NAT@parse\undefined
\makeatother
\usepackage[numbers,sort&compress]{natbib}

\usepackage[pdftex]{graphicx}
\usepackage{amsmath}
\usepackage{amssymb}  % assumes amsmath package installed
\usepackage{subcaption}
\usepackage{multirow}
\usepackage{array,booktabs}
\usepackage{diagbox}
\usepackage{balance}
\usepackage{xspace}
\usepackage{algorithm}
\usepackage{algpseudocode}
\usepackage{adjustbox}
\usepackage{romannum}
\usepackage[final]{hyperref}
\hypersetup{
 colorlinks=true,
 linkcolor=blue,
 filecolor=magenta,
 urlcolor=blue,
 citecolor=black
}
\usepackage{cleveref}
\crefname{figure}{Fig.}{Figs.}
\Crefname{figure}{Fig.}{Figs.}
\usepackage[font=small]{caption}
\usepackage{rpm_SIunits}
\usepackage{rpm_acronyms}
\usepackage{rpm_math}
\usepackage{rpm_misc}

\usepackage{soul,color}
\usepackage{lipsum}

\usepackage{xcolor}

\DeclareMathOperator*{\argmin}{argmin}

\usepackage{tikz} % \checkmark
\usepackage{caption}
\title{\LARGE \bf Ground-Optimized 4D Radar-Inertial Odometry \\ 
via Continuous Velocity Integration using Gaussian Process
}     

\author{Wooseong Yang${}^{1}$, Hyesu Jang${}^{2}$ and Ayoung Kim${}^{1*}$
\thanks{$^\dagger$This work was supported by the Robotics and AI (RAI) Institute, and in part by MOITE, Korea (No. 1415187329).}
\thanks{$^{1}$W. Yang and A. Kim are with the Dept. of Mechanical Engineering, SNU, Seoul, S. Korea {\tt\small [yellowish, ayoungk]@snu.ac.kr}}
\thanks{$^{2}$H. Jang is with the Institute of Advanced Machines and Design, SNU, Seoul, S. Korea {\tt\small pethesu@gmail.com}}
}
\begin{document}

%\onecolumn
\maketitle
\thispagestyle{empty}
\pagestyle{empty}

\begin{abstract}
% \textcolor{red}{Radar ensures robust sensing capabilities especially in adverse weather conditions, benchmarking as a promising sensor for robotics.
% Nevertheless, leveraging radars in robotics still poses challenges.
% The fundamental issue arises from the high level of noise. Intensity thresholding such as \ac{CFAR} has been employed, yet spatial uncertainties and generalizability issues persist.
% Another challenge is achieving accurate full 6-\ac{DOF} motion from doppler velocity with \ac{IMU}. Existing fusion strategy often exploited discretized propagation models with constant state assumptions, which are vulnerable to measurement errors.}

Radar ensures robust sensing capabilities in adverse weather conditions, yet challenges remain due to its high inherent noise level. Existing radar odometry has overcome these challenges with strategies such as filtering spurious points, exploiting Doppler velocity, or integrating with inertial measurements. This paper presents two novel improvements beyond the existing radar-inertial odometry: ground-optimized noise filtering and continuous velocity preintegration. Despite the widespread use of ground planes in LiDAR odometry, imprecise ground point distributions of radar measurements cause naive plane fitting to fail. Unlike plane fitting in LiDAR, we introduce a zone-based uncertainty-aware ground modeling specifically designed for radar. Secondly, we note that radar velocity measurements can be better combined with IMU for a more accurate preintegration in radar-inertial odometry.
Existing methods often ignore temporal discrepancies between radar and IMU by simplifying the complexities of asynchronous data streams with discretized propagation models.
Tackling this issue, we leverage GP and formulate a continuous preintegration method for tightly integrating 3-DOF linear velocity with IMU, facilitating full 6-DOF motion directly from the raw measurements.
Our approach demonstrates remarkable performance (less than 1\% vertical drift) in public datasets with meticulous conditions, illustrating substantial improvement in elevation accuracy. The code will be released as open source for the community: \textcolor{blue}{https://github.com/wooseongY/Go-RIO}.
\end{abstract}

% different temporal resolutions
% synch

%In this context, we propose ground-optimized radar-inertial odometry algorithm with continuous velocity integration.
%Since the ground can serve as a key feature in mitigating inherent noise, we introduce radar-specific ground modeling for reliable filtering.
% to mitigate imprecise ground distributions in radar.
% Moreover, utilizing weighted scan matching with velocity integration ensures robustness even with the measurement drifts.
%Secondly, even with an \ac{IMU}, achieving accurate 6-\ac{DOF} motion from radar reveals limitations arising from existing discretized models and constant state assumption. 
\section{Introduction}
\label{sec:intro}
%%%%%%%%%%%%%%%%%%%%%%%%%%%%%%%%%%%%%%%%%%%%%%%%%%%%%%%%%%%%%%%%%%%%%%
Robot navigation using radars has experienced remarkable advancements for their resilient perception capabilities in challenging weather conditions \cite{harlow2023new, abu2024radar, starr2014evaluation, bijelic2018benchmark, kramer2021radar}.
% Long electromagnetic wavelengths enable the radar to penetrate dust particles and detect objects over extended distances, even in low-visibility scenarios \cite{adams2012book, 9969174}.
Moreover, its capability of providing Doppler velocity distinguishes the radar sensor from the traditional sensors, leading to significant enhancements in visually degraded environments \cite{nissov2024degradation}.
% Despite these advancements, accurate motion estimation in radar navigation still presents significant challenges.

% vertical limitation

Despite the above advantages, radar measurements are vulnerable to false alarms and multipath reflections, thus necessitating outlier rejection.
Among many existing methods, intensity-based filtering \cite{adolfsson2023lidar} and \ac{CFAR} are the most widely utilized. However, these approaches are threshold-sensitive and can be limited in terms of generalizability.
In this context, leveraging ground could be a feasible solution as it demonstrated its reliable noise filtering ability in \ac{LiDAR} \cite{shan2018lego, pan2021mulls, zheng2021efficient, chen2021low, wei2022gclo, galeote2023gnd}. Unfortunately, naive plane fitting as in \ac{LiDAR} would fail because of the inherent uncertainty and imprecise distribution in radar \cite{li20234d, chen2023drio}.
% , which highlights the necessity for radar-specific ground modeling.
% Regarding these issues, leveraging ground plane has actively exploited as a consistent feature in \ac{LiDAR} \cite{shan2018lego, pan2021mulls, zheng2021efficient, chen2021low, wei2022gclo, galeote2023gnd}.
% Unlike \ac{LiDAR}, ranges from radar involve noises and spatial uncertainties and represent imprecise distributions of ground points, which poses significant challenge \cite{chen2023drio}.
% Moreover, single-plane model \cite{li20234d} relies on ground classification through height thresholding, which struggles with the complexities of slopes and hilly terrains.
% These necessitate radar-tailored ground segmentation.
% Our piece-wise ground extraction incorporating point-wise spatial uncertainties can resolve the aforementioned issues.

While mitigating inherent noise in radar, estimating 6-\ac{DOF} motion from the 3-\ac{DOF} linear velocity presents additional challenges.
Over the past decades, ego-motion estimation using Doppler velocity has been explored \cite{kellner2013instantaneous, kramer2020radar, haggag2022credible}, and has been extensively advanced to odometry by integrating \ac{IMU} \cite{DoerENC2020, park20213d, michalczyk2022tightly, zhuang20234d, do2024dero}.
In doing so, unfortunately, existing methods overlooked the temporal discrepancies from two sensors, radar and \ac{IMU}, and the temporal discrepancy in sensor data streams prohibits the tight integration of two sensors for accurate preintegration.
\begin{figure}[!t]
    \centering
    \includegraphics[trim=9cm 0.1cm 7.8cm 0cm, clip, width=0.99\columnwidth]{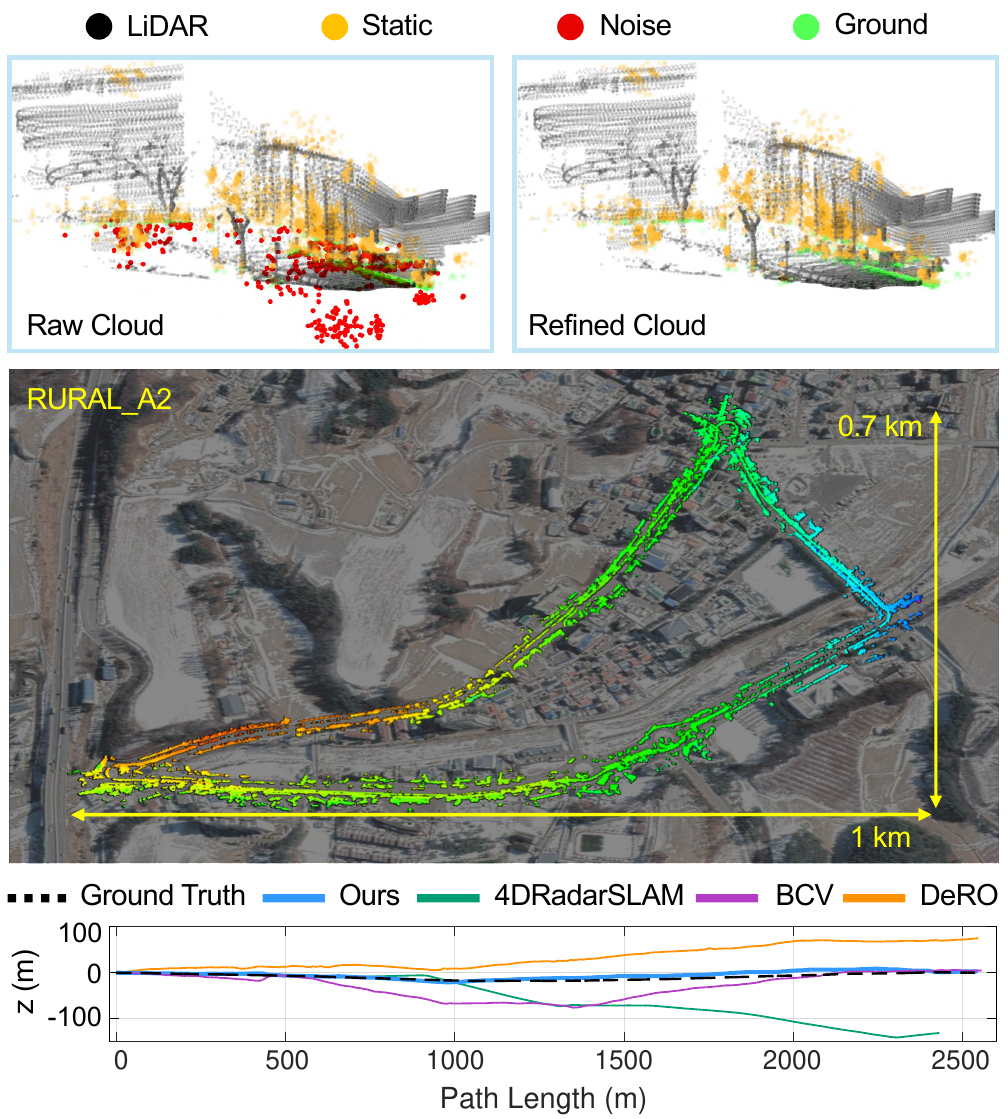}
    \caption{
        \textbf{Top}: Our uncertainty-aware ground filtering efficiently eliminates the noise (red) from radar.
        \textbf{Middle}: Mapping result of \texttt{RURAL\_A2}. Our continuous velocity integration with the Gaussian Process (GP) proficiently handles the sharp turns and roundabouts.
        \textbf{Bottom}: The proposed method shows the lowest elevation error (18m over 2.5km of path length, only \textbf{0.72\%}).
    }
    \label{fig:overview}
    \vspace{-8mm}
\end{figure}
%FIGURE

% Recently, incorporating spatial information has proposed \cite{michalczyk2022tightly, 10160482, zhuang20234d, zhang20234dradarslam, do2024dero}.
% Conventional filtering methods that rely on distance or signal intensity \cite{kung2021normal, michalczyk2022tightly, 10160482, zhuang20234d} tend to be parameter-sensitive and lack generalizability. Filtering with RANSAC \cite{zhang20234dradarslam, do2024dero} introduces measurement drift for large objects \cite{CasadoHerraez2024radar}. Particularly, radar's low resolution along the elevation and inherent measurement uncertainties exacerbate these issues.
% These difficulties necessitate further advancements in filtering techniques in 4D radar that can effectively embed the complexities and uncertainties in radar measurement.

\begin{figure}[!t]
    \centering
    \includegraphics[trim=3.7cm 4.5cm 7cm 0.2cm, clip, width=0.95\columnwidth]{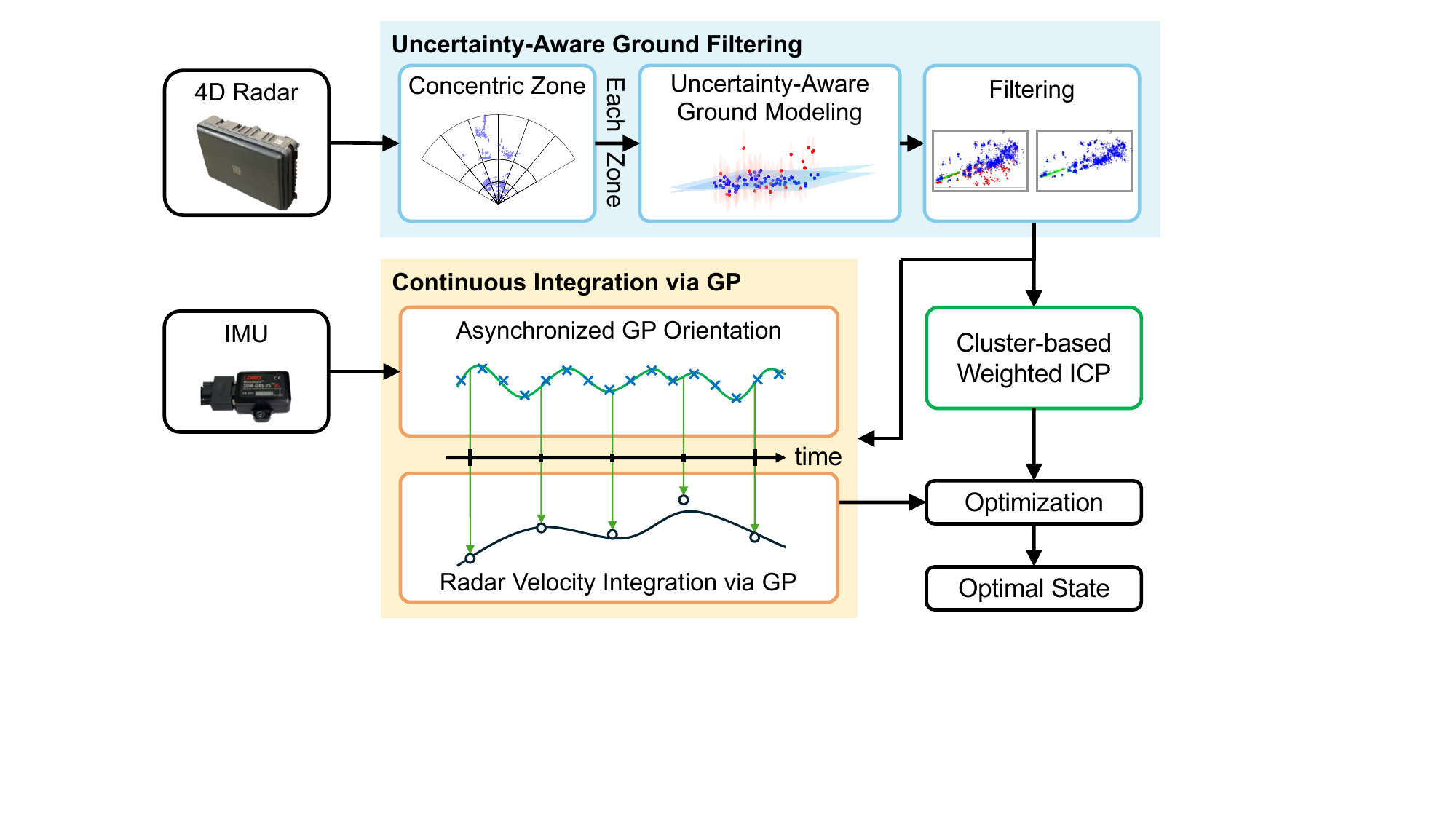}
    \caption{
        The overall pipeline of our algorithm.
    }
    \label{fig:pipeline}
    \vspace{-7mm}
\end{figure}

In this paper, we propose a tightly-coupled 4D radar-inertial odometry framework with ground filtering and continuous preintegration for radar-inertial fusion. Existing ground models utilized height thresholding, which struggled with the complexities of slopes and hills. Instead, we use a zone-based approach incorporating the spatial uncertainties from the radar point cloud into ground modeling.
Also, radar-inertial fusion is performed through continuous preintegration of velocities.
Inspired by \ac{IMU} preintegration using \ac{GP} \cite{le2021continuous}, we found that formulating the radar velocities into \ac{GP} allows seamless integration with \ac{IMU} and effectively addresses temporal discrepancies that are often overlooked in existing works.
% allowing for robust 6-\ac{DOF} motion estimation directly from the asynchronous measurements from radar and IMU.
\figref{fig:overview} and \figref{fig:pipeline} illustrates the improvements and the pipeline of our method.
The key contributions are as follows:

\begin{itemize}
    \item \textbf{Uncertainty-Aware Radar Ground Filtering: }
    Leveraging ground in radar should overcome the ambiguous distribution of the radar point cloud. By associating spatial uncertainties with zone-based ground modeling, we effectively segment out the ground even with its imprecise distributions and secure reliable noise filtering.

    % Identifying the ground points from the radar poses significant challenge due to distribution ambiguity. Our ground model captures the spatial uncertainty arises from radar, effective handling the inherent noises.
    
    \item \textbf{Novel Continuous Radar Velocity Preintegration: }
    Effectively mitigating the temporal discrepancies between radar and \ac{IMU} often disregarded in existing methods. We reformulate radar velocities and \ac{IMU} using \ac{GP} for the tightly-coupled preintegration, enabling robust motion estimation directly from the asynchronous measurements without any assumption.
    % integration problem / emphasize GP's proficient
    % novel continuous preintegration of radar velocities and imu using GP which enabling assumption-free motion estimation directly from the measurments
    
    \item \textbf{Evaluation on Diverse Environmental Conditions: }
    The proposed method is thoroughly validated on real-world public datasets including high dynamics and harsh weathers, demonstrating robust performance compared to existing 4D radar-inertial odometry algorithms. In particular, our approach achieves substantially enhanced odometry performance in the vertical direction.

    % various evaluation and achieve enhanced odometry in $z$-axis
\end{itemize}
\section{related work}
\label{sec:relatedwork}

\subsection{Ground-Aided LiDAR and Radar Odometry}

The ground is a widely exploited feature for alleviating vertical drift in \ac{LiDAR} odometry owing to stability and consistency.
LeGO-LOAM \cite{shan2018lego} pioneered the utilization of the ground plane as a planar feature to estimate roll, pitch, and $z$-axis motion. MULLS \cite{pan2021mulls} integrated the ground plane as a geometric feature within multi-metric matching. \citet{zheng2021efficient} leveraged the ground plane for decoupled registration.
\citet{chen2021low} employed ground segmentation to eliminate redundant features and registration through semantic labeling.
GCLO \cite{wei2022gclo} utilized the ground as a planar landmark to mitigate vertical drift. GND-LO \cite{galeote2023gnd} achieved 2D motion with planar patches from the ground.

The focus on utilizing ground features in LiDAR odometry has been further extended to radar. \citet{li20234d} employed a RANSAC-based ground plane model for noise filtering. DRIO \cite{chen2023drio} optimized ego-velocity using ground points in single-chip radar. However, these methods relied on heuristic height thresholds to identify ground points, which can be problematic in sloped terrains.
% Furthermore, naive ground fitting with the single-plane model is problematic in sloped terrains.
To overcome these challenges in radar ground detection, our system incorporates spatial uncertainties from radar with zone-based ground modeling.

\subsection{Radar-Inertial Odometry}

Similar to visual and \ac{LiDAR} odometry, fusion with \ac{IMU} has shown promising advancements. EKF-RIO \cite{DoerENC2020} fused radar and \ac{IMU} using \ac{EKF} with online calibration, however, the assumption of synchronized measurements was the limitation. \citet{ng2021continuous} proposed a continuous trajectory model on $SE(3)$ to mitigate temporal discrepancies between radar and \ac{IMU}, yet it encountered $z$-axis drifts due to high uncertainties on elevation. To address these limitations, \citet{park20213d} configured two orthogonal radars to achieve accurate 3D velocity with a velocity factor.

Recent works have integrated spatial information in odometry. \citet{michalczyk2022tightly} proposed \ac{EKF}-based radar-inertial odometry with Stochastic Cloning.
4D-iRIOM \cite{zhuang20234d} leveraged graduated non-convexity for ego-velocity estimation with scan-to-submap matching via \ac{EKF}.
DeRO \cite{do2024dero} utilized the radar velocity in the propagation model with scan matching.
Despite these advances, temporal discrepancies between radar and \ac{IMU} are still overlooked, and unresolved challenges remain concerning the vertical drifts. 
To tackle these issues, we introduce a continuous radar-inertial fusion framework with \ac{GP}, which enables assumption-free motion estimation directly from the asynchronous measurements.
% \citet{huang2024morephysicalenhancedradarinertialodometry} leveraged a Radar Cross Section (RCS) bounded filter for refining radar point-wise correspondences.
\section{Methodology}
\label{sec:method}

%FIGURE
%FIGURE

%In this section, we introduce the overall system of our proposed algorithm. The pipeline is shown in \figref{fig:pipeline}, and detailed descriptions are provided in the following subsections.

\subsection{Uncertainty-Aware Radar Ground Filtering}
\label{subsec: ground segmentation}
% ground plane detection and outlier removal

\begin{figure}[!t]
    \centering
    % \hspace{-1.5cm}
    \begin{subfigure}[b]{0.45\columnwidth}
        \includegraphics[trim=0cm 12cm 23cm 0cm, clip,width=\textwidth]{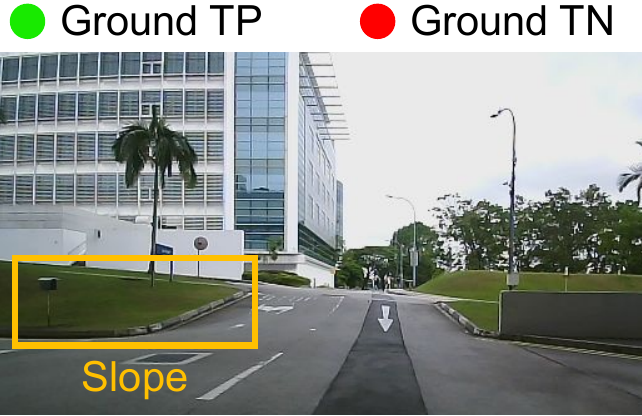}
        \caption{Scene including a slope}
        \label{fig:hill_a}
    \end{subfigure}
    \begin{subfigure}[b]{0.45\columnwidth}
        \includegraphics[trim=0cm 0cm 0cm 0cm, clip,width=\textwidth]{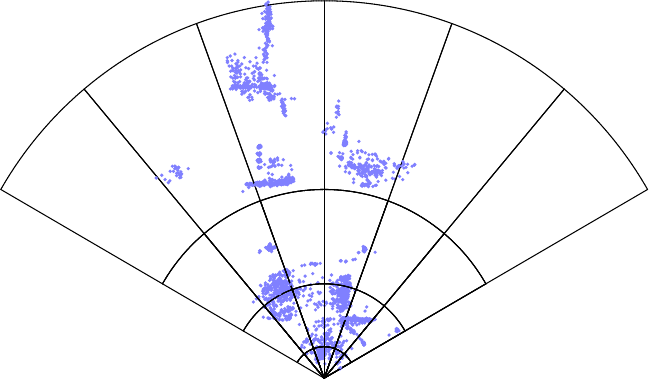}
        \caption{Concentric Zone model}
        \label{fig:hill_b}
    \end{subfigure}
    \\
    % \hspace{-0.7cm}
    \begin{subfigure}[b]{0.45\columnwidth}
        \includegraphics[trim=1cm 12cm 24cm 1.2cm, clip,width=\textwidth]{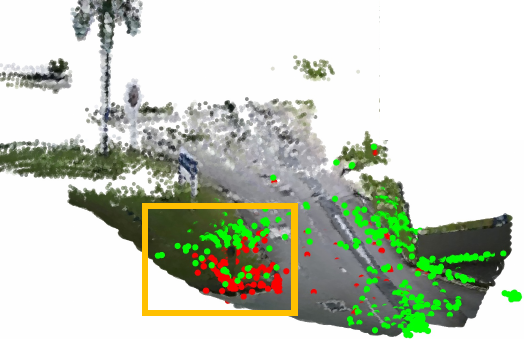}
        \caption{Naive plane fitting}
        \label{fig:hill_c}
    \end{subfigure}
    \begin{subfigure}[b]{0.45\columnwidth}
        \includegraphics[trim=1cm 12cm 24cm 1.2cm, clip,width=\textwidth]{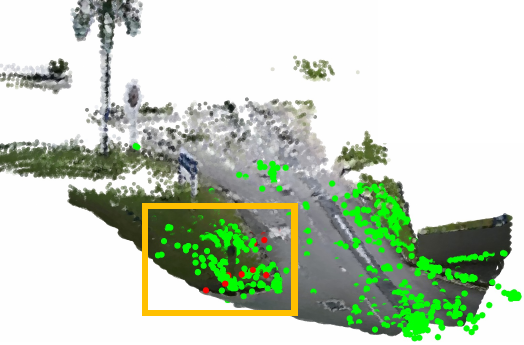}
        \caption{Proposed zone-based fitting}
        \label{fig:hill_d}
    \end{subfigure}
    %\vspace{-0.2cm}
    \caption{
    Comparison between naive plane fitting and our zone-based approach in the sloped environment (yellow box). (c) Many true negatives are found when using naive plane fitting (TN, red). (d) The proposed model effectively filtered ground points even at slope as true positives (TP, green).}
    \label{fig:ground}
    \vspace{-7mm}
\end{figure}

Radar points are prone to degradation due to noise from false alarms and multipath effects, necessitating noise filtering to ensure robustness.
We primarily employ radius filtering to eliminate clutter points, and the ground is utilized to mitigate the multipath.
To overcome the issues arising from slopes and hills in the existing naive ground fitting, zone-based ground detection is proposed.
Considering the conical shape of the radar point cloud, we adopt \ac{CZM} \cite{lee2022patchworkpp}, which is characterized by increasing width with distance as depicted in \figref{fig:ground}.

% This model features an increasing width with distance, ensuring equitable distribution of point densities across regions compared to standard uniform voxelization.

The traditional approach for estimating plane models exploited \ac{PCA}. Despite their prevalent usage, \ac{PCA} presents limitations regarding sensitivity to noise and uncertainties, which are crucial for radars.
To resolve this, we formulate the plane estimation as an optimization problem that minimizes the sum of Mahalanobis distances between the points and the plane.
The Mahalanobis distance from the point $p_i$ to plane is computed as $D_{M_i} = \sqrt{(\mathbf{n}^\top+d)^\top \Sigma_i^{-1} (\mathbf{n}^\top+d)}$, while $\Sigma_i$ is the associated point-wise covariance matrix, and $[\textbf{n},d]$ represents the detected plane model. Then the plane model can be estimated by the subsequent optimization:
% The point-wise covariance matrix $\Sigma_i$ of the point $p_i$ can be computed \cite{zhang20234dradarslam}. 
\vspace{-2mm}
\begin{equation}
\begin{array}{rrclcl}
    \displaystyle \argmin_{[\textbf{n},d]} & \multicolumn{3}{l}{\sum_{i=1}^n D_{M_i}^2} \\
    % \vspace{-2mm}
    % \textbf{v}_i &=\textbf{n}^\top p_i+d \\
    \kappa &= \textbf{n}^\top C\textbf{n}.
    % \vspace{-1mm}
\end{array}
\label{eqn:plane_model}
\end{equation}
$C\in\mathbb{R}^{3\times 3}$ is the covariance matrix of the point cloud considering spatial uncertainty, and $\kappa$ means flatness.
We merge the points adjacent to the estimated plane and refine the ground set by excluding points that exhibit significant deviations from the plane. These are iterated until the flatness converges in each divided region, and the points below the ground are classified as noise and removed. The entire ground segmentation process is described in Algorithm~\ref{alg:ground}.

% ALGORITHM
\begin{algorithm}[!b]
   \begin{algorithmic}[1]
   \footnotesize
    \caption{Uncertainty-aware Ground Segmentation}
    \label{alg:ground}
    \State  \textbf{Input:} PointCloud $\mathcal{P}$, sensor height $h_s$, $\epsilon_d, \epsilon_f$; 
    \State  \textbf{Output:} Ground PointCloud $\mathcal{G}$, Static PointCloud $\mathcal{P}^r$;
    \State \% Divide the pointcloud into each zone.
    \State $\mathcal{P} \ = $ \{$\mathcal{P}_1$, $\cdots$, $\mathcal{P}_{N_z}$\}, $\mathcal{G}=\emptyset, \mathcal{P}^r=\emptyset$;
    \For{$i=1,\cdots,N_z$}
        \If{size$(P_i)>20$}
        \State \% initialize
        \State $\mathcal{G}_i = \{p_i\in \mathcal{P}_i, \text{ s.t. } z_i<-h_s/2\}, \mathcal{Q}_i=\mathcal{P}_i-\mathcal{G}_i$;
        \Repeat
        \State estimate the ground plane with $\mathcal{G}_i$ using \eqref{eqn:plane_model};
        \State $\mathcal{G}_i \leftarrow \mathcal{G}_i \cup \{ p_i\in \mathcal{Q}_i, \text{ s.t. } D_{M_i}<\epsilon_d \}$;
        \State $\mathcal{Q}_i \leftarrow \mathcal{Q}_i - \{ p_i\in \mathcal{Q}_i, \text{ s.t. } D_{M_i}<\epsilon_d \}$;
        \State $C_i =$ \textit{computeCovariance}$(\mathcal{G}_i)$;
        \Until \textit{flatness}$(\mathcal{G}_i)$ $<\epsilon_f$
        \State $\mathcal{Q}_i\leftarrow\mathcal{Q}_i-\{ p_i\in \mathcal{Q}_i, \text{ s.t. below the ground} \}$;
        \State $\mathcal{G}\leftarrow\mathcal{G}\cup\mathcal{G}_i, \mathcal{P}^r\leftarrow\mathcal{P}^r\cup\mathcal{Q}_i$;
        \EndIf
    \EndFor
    \State \textbf{return} $\mathcal{G}, \mathcal{P}^r$;
    \normalsize
    \end{algorithmic}
\end{algorithm}
% ALGORITHM

After segmentation, the refinement of the height value $z_i$ of the point $p_i=
(x_i,y_i,z_i)$ from the ground is performed with the following equation derived from point-wise radial velocity measurement ${v_d}_i$ \cite{DoerENC2020}:
\vspace{-2mm}
\begin{equation}
    z_i=\cfrac{V_{xy}v_z - \sqrt{V_{xy}^2{{v_d}_i}^2 - ({{v_d}_i}^2-v_z^2)(x_i^2+y_i^2){{v_d}_i}^2}}{{{v_d}_i}^2-v_z^2},
    \vspace{-2mm}
\end{equation}
where $v_x, v_y, v_z$ is the vehicle's ego velocity, and $V_{xy}=v_x x_i+v_y y_i$. The dynamic objects on non-ground points are removed after the refinement process.

\subsection{Continuous Velocity Preintegration via Gaussian Process}
\label{subsec:velocity integration}

Our system leverages a \ac{GP}-based continuous modeling on radar velocities, seamlessly integrating 4D radar velocity and \ac{IMU}.
Existing radar-inertial odometry approaches have assumed synchronized measurements \cite{DoerENC2020} or employed the discretized propagation model \cite{zhuang20234d} to mitigate temporal discrepancies.
Inspired by \citet{le2021continuous}, who overcome the above limitations in \ac{IMU} by introducing \ac{GP}, our work leverages \ac{GP} for tightly-coupled preintegration of radar velocity and \ac{IMU}. The continuous nature of the \ac{GP} resolves the temporal discrepancies and enables direct motion estimation without any assumptions despite the asynchronous data streams between radar velocity and \ac{IMU}.

Let us briefly summarize \cite{le2021continuous} and introduce our radar-\ac{IMU} preintegration formulation.
%The angular velocities reflect the orientation change rate, represented in the tangent space of the $SO(3)$ manifold. Since the time derivative of the rotation vector $\boldsymbol{\theta}\in\mathbb{R}^3$ captures this change in orientation, the angular velocity is achieved by the chain rule and its right Jacobian $J_r(\boldsymbol{\theta})$:
%\vspace{-1mm}
%\begin{equation}
%    \boldsymbol{\omega}(t) = J_r(\boldsymbol{\theta}(t))\dot{\boldsymbol{\theta}}(t).
%\end{equation}
%
% To obtain orientation from the angular velocity measured by the \ac{IMU}, preintegration \cite{forster2017manifold} is commonly employed. However, this approach exploited a discretized propagation model with the constant states assumptions, aggravating errors from measurement drifts. To tackle this issue, we utilize \ac{GP} regression, which offers significant advantages in continuous inference and integration \cite{sarkka2011linear, le2021continuous}.
%
Each component of the time derivative of the rotation vector $\dot{\boldsymbol{\theta}}=(\dot\theta_x, \dot\theta_y, \dot\theta_z)$ is modeled using a zero-mean \ac{GP}:
\vspace{-1mm}
\begin{equation}
    \dot\theta_i(t)\sim\mathcal{GP}(0,k_{\theta_i}(t,t')),
\end{equation}
where $k_{\theta_i}(t,t')$ is the kernel function and the index $i$ represents $x,y,z$ components each.
Then the inference ($\star$) of the rotation vector is conducted as follows:
\begin{eqnarray}
\begin{aligned}
\label{eqn:rot_derivative_inference}
    \dot{\theta}_{i\star}(t)&=\mathbf{k}_{\theta_i}(t,\mathbf{t})(\mathbf{K}_{\theta_i}(\mathbf{t}, \mathbf{t})+\sigma_i^2\mathbf{I})^{-1}\boldsymbol{\rho}_i \\
% \end{equation}
% \begin{equation}
\label{eqn:rot_inference}
    \theta_{i\star}(t)&=\int\mathbf{k}_{\theta_i}(t,\mathbf{t})(\mathbf{K}_{\theta_i}(\mathbf{t}, \mathbf{t})+\sigma_i^2\mathbf{I})^{-1}\boldsymbol{\rho}_i \partial t \\
% \end{equation}
% \begin{equation}
\label{eqn:kernel}
    \mathbf{k}_{\theta_i}(t,\mathbf{t})&=[k_{\theta_i}(t,t_1)\dots k_{\theta_i}(t,t_{N_i})] \\
    \mathbf{K}_{\theta_i}(\mathbf{t},\mathbf{t})&=
    \begin{bmatrix}
        k_{\theta_i}(t_1,t_1) & \cdots & k_{\theta_i}(t_1,t_{N_i}) \\
        \vdots & \ddots & \vdots \\
        k_{\theta_i}(t_{N_i},t_1) & \cdots & k_{\theta_i}(t_{N_i},t_{N_i})
    \end{bmatrix},
\end{aligned}
\end{eqnarray}
$\mathbf{t}=[t_1\dots t_{N_i}]$ is the vector of the \ac{IMU} measurement timestamps,
$\mathbf{k}_{\theta_i}(t,\mathbf{t})$ is the kernel vector.
$\sigma_i$ is the standard deviation of the \ac{IMU} gyroscope, and $\boldsymbol{\rho}_i$ is the vector of $\hat{\dot{\theta}}_i(\mathbf{t})$, which can be achieved with the following optimization:
\vspace{-1mm}
\begin{equation}
    \displaystyle \argmin_{[\boldsymbol{\rho}_x, \boldsymbol{\rho}_y, \boldsymbol{\rho}_z]}\sum_{n=1}^{N_i} \left( \norm{\boldsymbol{e}_n^{meas}}_{\Sigma_{\boldsymbol{\omega}}}^2+\norm{\boldsymbol{e}_n^{gp}}_{\Sigma_{gp}}^2 \right),
\end{equation}
where $\norm\cdot_\Sigma$ represents the Mahalanobis norm.
The first term, $\boldsymbol{e}_n^{meas} = J_r(\boldsymbol{\theta}_\star(t_n))\dot{\boldsymbol{\theta}}_\star(t_n)-\Tilde{\boldsymbol{\omega}}(t_n)$ is about the measurement constraint, and $\Sigma_{\boldsymbol{\omega}}$ is the covariance matrix of \ac{IMU} gyroscope. The angular velocity is achieved by the chain rule and its right Jacobian $J_r(\boldsymbol{\theta})$ as $\boldsymbol{\omega}(t) = J_r(\boldsymbol{\theta}(t))\dot{\boldsymbol{\theta}}(t)$.
Secondly, $\boldsymbol{e}_n^{gp}=\dot{\boldsymbol{\theta}}_{\star}(t_n)-\hat{\dot{\boldsymbol{\theta}}}(t_n)$ constraints the unique solution of the optimization with $\Sigma_{gp}$ being \ac{GP} variance.
%These optimized inducing values enable inference of the rotational increment $\Delta \mathbf{R}$ at any given time.

The key idea is that when formulating \ac{GP} to tightly couple radar and \ac{IMU}, \eqref{eqn:rot_inference} enables the inference of rotation at any given time, allowing radar-\ac{IMU} velocity preintegration without additional time synchronization.
As the velocity $\mathbf{v}=(v_x,v_y,v_z)$ cannot be modeled with a zero-mean \ac{GP}, we conduct \ac{GP} model with the prior mean function $\boldsymbol{\mu}(t)$:
\vspace{-1mm}
\begin{equation}
    {v}_i(t)\sim\mathcal{GP}(\mu_i(t),k_{v_i}(t,t')).
\end{equation}

Then the inference of the velocity and translation can be performed as following:
\vspace{-1mm}
\begin{equation}
    {v}_{i\star}(t)=\mu_i(t)+\mathbf{k}_{v_i}(t,\mathbf{t})(\mathbf{K}_{v_i}(\mathbf{t}, \mathbf{t})+\epsilon_i^2\mathbf{I})^{-1}(\boldsymbol{\zeta}_i-\mu_i(\mathbf{t}))
\end{equation}
% \vspace{-1mm}
\begin{equation}
\label{eqn:trans_inference}
    {p}_{i\star}(t)=\int \mu_i(t)+\mathbf{k}_{v_i}(t,\mathbf{t})(\mathbf{K}_{v_i}(\mathbf{t}, \mathbf{t})+\epsilon_i^2\mathbf{I})^{-1}(\boldsymbol{\zeta}_i-\mu_i(\mathbf{t})) \partial t,
\end{equation}
where $\boldsymbol{\zeta}_i$ is the vector of $\hat{v}_i(\mathbf{t})$, $\mathbf{t}=[t_1\dots t_{N_r}]$ represents the radar timestamps, and $\epsilon_i$ is the standard deviation of the velocity measurements.
Hence, inducing values $\boldsymbol{\zeta}_i$ are achieved by the following optimization:
\vspace{-1mm}
\begin{equation}
\label{eqn:vel_optimize}
    \displaystyle \argmin_{[\boldsymbol{\zeta}_x, \boldsymbol{\zeta}_y, \boldsymbol{\zeta}_z]}\sum_{n=1}^{N_r} \left( \norm{\boldsymbol{e}_n^{meas}}_{\Sigma_{\mathbf{v}_n}}^2+\norm{\boldsymbol{e}_n^{gp}}_{\Sigma_{gp}}^2 \right).
\end{equation}
Now we have the radar-associated residual as \eqref{eqn:vel_optimize} with two residual terms: $\boldsymbol{e}_n^{meas} = \Delta{\mathbf{R}^\top}\mathbf{v}_\star(t_n)-\Tilde{\mathbf{v}}(t_n)$, and $\boldsymbol{e}_n^{gp}=\mathbf{v}_{\star}(t_n)-\hat{\mathbf{v}}(t_n)$. Here, $\Sigma_{\mathbf{v}_n}$ is the covariance matrix of velocity at timestamp $t_n$. After \eqref{eqn:vel_optimize} has converged, motion increment $\textbf{T}^{\text{INT}}\in SE(3)$ can be inferenced by \eqref{eqn:rot_inference} and \eqref{eqn:trans_inference}.

\subsection{Cluster-based Weighted ICP}
\label{subsec:scan matching}
% Radar provides spatial information, we utilize scan matching to determine the transformation between consecutive keyframes. In APD-GICP \cite{zhang20234dradarslam}, both point-wise covariance and point-to-plane error in radar scans are considered, addressing the previously neglected matching quality of correspondences. To further enhance the stability of these correspondences, we introduce a weighting mechanism to the objective function in APD-GICP.

Although individual radar points exhibit inherent uncertainties, clusters from radar scans can effectively represent the overall shape of structures as shown in \figref{fig:cluster}. These clusters often correspond to prominent features, which are crucial for accurate scan registration. We utilize the DBSCAN to segment these clusters and assign higher weights to the correspondences within consistent clusters. Furthermore, the flatness of each cluster is incorporated into the weight calculation, enhancing the spatial representation of each cluster.
Therefore, the transformation $\textbf{T}^{\text{ICP}}\in SE(3)$ between keyframes is estimated by the following optimization:
\begin{equation}
    \displaystyle \textbf{T}^{\text{ICP}}=\argmin_{\textbf{T}} \sum_i \left(w_{i, clust}+w_{i,flat}\right)\norm{p_i-\textbf{T}q_i}_{\Sigma_i}^2
\end{equation}
where $\Sigma_i$ is the covariance matrix including spatial uncertainty, and $w_i$ represents the each weight.

% This approach enhances the robustness and accuracy of scan matching by leveraging the structural information inherent in radar point clouds and by adjusting the influence of correspondences based on their spatial relationships and geometric properties.

\subsection{Pose Graph Optimization}

Incremental motion from \secref{subsec:velocity integration} and \secref{subsec:scan matching} is used to define $SE(3)$ edges in the pose graph, and the keyframe-based pose graph optimization is conducted as:
\vspace{-1mm}
\begin{eqnarray}
\label{eqn:pgo}
    \argmin_{\textbf{X}}\sum_{(i,j)}\norm{\textbf{T}^{\text{INT}}_{ij}-h(\textbf{x}_i,\textbf{x}_j)}^2_{\Sigma_\text{INT}}+\norm{\textbf{T}^{\text{ICP}}_{ij}-h(\textbf{x}_i,\textbf{x}_j)}^2_{\Sigma_\text{ICP}}
\end{eqnarray}
where $h(\textbf{x}_i,\textbf{x}_j)$ computes the relative transformation between two poses. 
A new keyframe is selected when the motion increment from numerical integration exceeds a predefined threshold relative to the previous keyframe. The $g2o$ \cite{kummerle2011g} library is utilized for back-end optimization.

%FIGURE
\begin{figure}[!t]
    \centering
    \includegraphics[trim=0cm 1cm 9.5cm 0cm, clip, width=1\columnwidth]{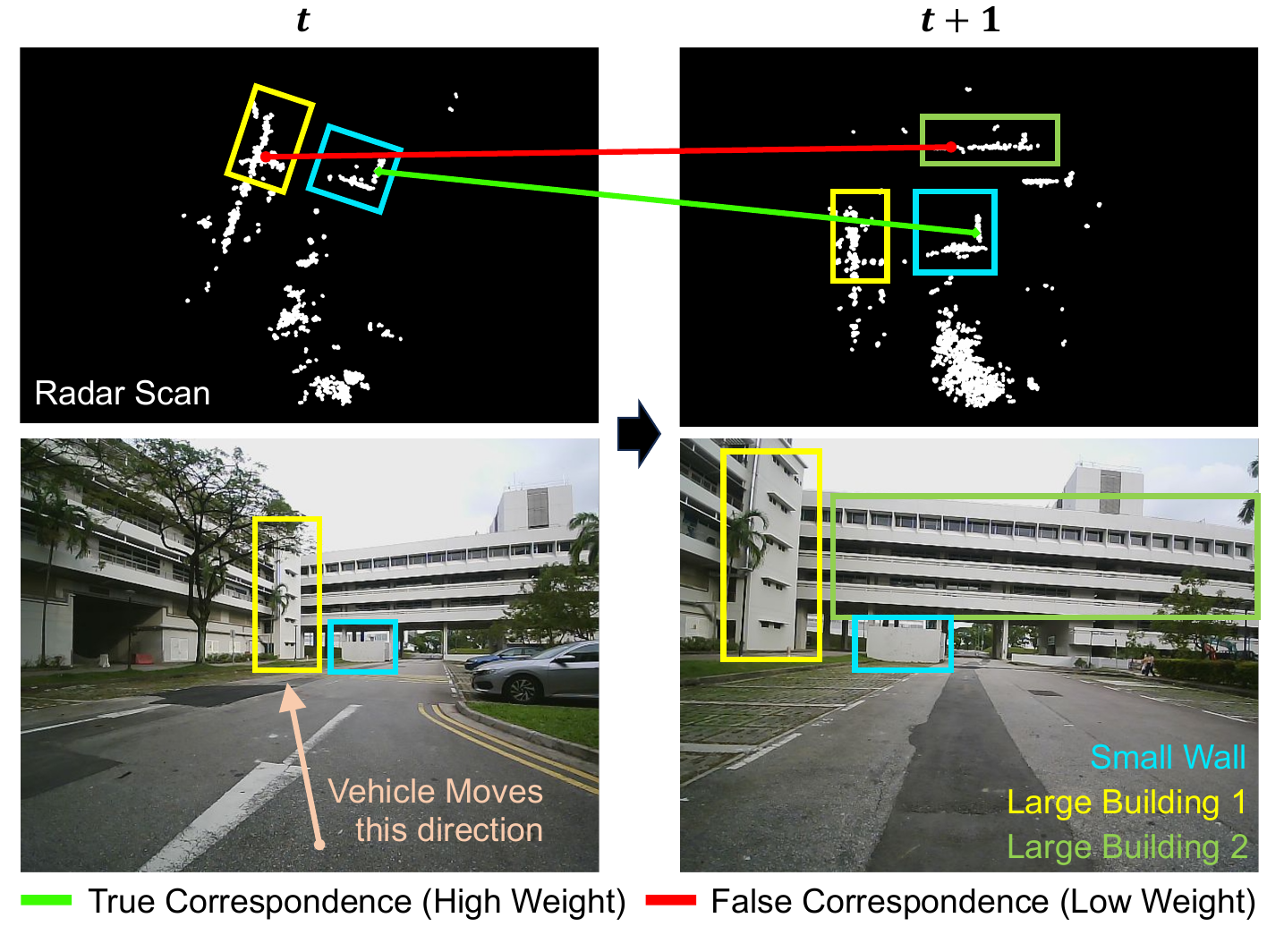}
    \vspace{-7mm}
    \caption{
        As the vehicle moves $t$ to $t+1$, prominent features maintain the structures in radar (yellow, cyan). The higher weight is allocated to correspondences within the consistent cluster (green line).
    }
    \label{fig:cluster}
    \vspace{-6mm}
\end{figure}
%FIGURE
\section{experiment}
\label{sec:experiment}

\subsection{Datasets and Evaluation Metric}

For evaluation, we use NTU4DRadLM \cite{zhang2023NTU} and MSC-RAD4R \cite{choi2023MSC} datasets, both utilized Oculii Eagle 4D radar and \ac{IMU}. NTU4DRadLM is designed to assess performance in semi-structured environments with ground vehicle systems. MSC-RAD4R includes urban and rural environments using a car platform with fast and high-dynamic sequences.
% For keyframe selection, we use 1\unit{m} and 5{\textdegree} for thresholds in handcart sequences, and 5\unit{m} and 10{\textdegree} for car sequences.

We benchmark our method against state-of-the-art open-sourced 4D radar and radar-inertial odometry algorithms: APD-GICP in 4DRadarSLAM \cite{zhang20234dradarslam}, EKF-RIO \cite{DoerENC2020} BCV \cite{park20213d}, and DeRO \cite{do2024dero}.
Unfortunately, other ground-aided radar-inertial odometry \cite{chen2023drio, li20234d} are not open-sourced for comparison. Instead, a thorough comparative analysis for our uncertainty-aware ground filtering will be provided in \secref{subsec:ablation_ground}. We remove the loop closure in 4DRadarSLAM to ensure equitable comparison, as our approach focuses solely on odometry. For quantitative analysis, we compute the \ac{RMSE} of the \ac{ATE} and \ac{RPE} with evo library \cite{grupp2017evo}.
\ac{ATE} is expressed in meters, and \ac{RPE} is expressed in degrees per meter and percentage for the rotational (RPE$_r$) and translational (RPE$_t$) components, respectively.
Since MSC-RAD4R provides ground truth data only for translation, we evaluated only the \ac{ATE} in MSC-RAD4R.
In each table, we highlight the best results in \textbf{bold}.

\subsection{Evaluation on the NTU4DRadLM Dataset}
\label{subsec:ntu}

%TABLE
\begin{table*}[t!]
\scriptsize
\label{table:ATE}
\centering
\caption{ATE (m) Evaluation Results in NTU4DRadLM and MSC-RAD4R (Best results in \textbf{bold})}
\vspace{-1mm}
\label{tab:result_ate}
\resizebox{1\linewidth}{!}{
\begin{tabular}{l|cccc|cccccccc}
\hline
\textbf{Method} & \textbf{cp} & \textbf{nyl} & \textbf{loop2} & \textbf{loop3} & \textbf{URBAN\_A0} & \textbf{LOOP\_A0} & \textbf{LOOP\_B0} & \textbf{LOOP\_C0} & \textbf{LOOP\_D0} & \textbf{LOOP\_E0} & \textbf{RURAL\_A2} & \textbf{RURAL\_B2} \\ \hline\hline
4DRadarSLAM \cite{zhang20234dradarslam} & 0.768 & 7.574 & 131.948 & \textbf{26.755} & 16.097 & 110.269 & 88.228 & 14.333 & 19.984 & 56.347 & 33.256 & 24.251 \\ 
EKF-RIO\cite{DoerENC2020}               & 6.241 & 58.913 & 436.262 & 181.422 & 18.638 & 96.662 & 159.662 &176.106 & 26.895 & 196.652 & 278.788 & 251.471 \\  
BCV\cite{park20213d}                    & 3.163 & 41.002 & 222.182 & 231.449 & 2.667 & 120.297 & 130.256 & 81.444 & 75.382 & 95.018 & 19.423 & 14.256 \\ 
DeRO\cite{do2024dero}                   & 4.910 & 48.275 & 438.643 & 146.659 & 12.317 & 149.277 & 39.620 & 51.058 & 46.708 & 186.376 & 263.003 & 73.784 \\ 
Proposed                                & \textbf{0.699} & \textbf{5.009} & \textbf{42.185} & 28.269 & \textbf{1.757} & \textbf{16.752} & \textbf{9.884} & \textbf{6.450} & \textbf{4.280} & \textbf{35.266} & \textbf{9.480} & \textbf{10.927}\\  \hline
\end{tabular}}
\vspace{-2mm}
\end{table*}
%TABLE

\begin{figure*}[!t]    
    \centering
    % Left Minipage
    % \hspace{1mm}
    \begin{minipage}{0.49\textwidth}
        \vspace{-70mm}
        \begin{subfigure}[b]{1.0\textwidth}
            \centering
            \hspace{-5mm}
            \includegraphics[trim=2.7cm 0cm 6.7cm 0.6cm, clip, width=1\textwidth]{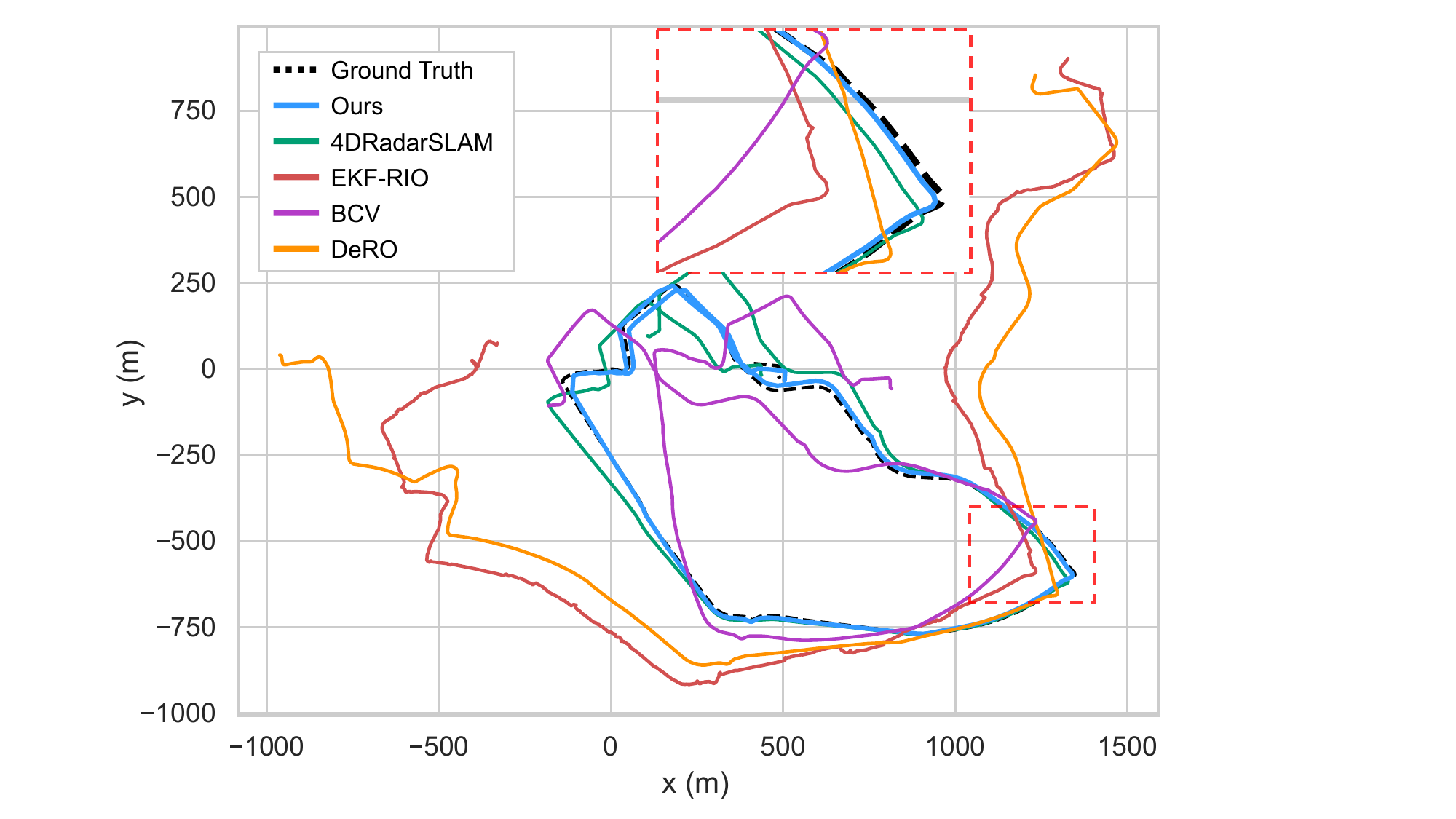}
            \vspace{-3mm}
            \caption{Estimated trajectory in \texttt{loop2}}
            \vspace{-4mm}
            \label{subfig:ntu_loop2_traj}
            \setcounter{figure}{5}
        \end{subfigure}
    \end{minipage}
    % Right Minipage
    \hspace{-2mm}
    \begin{minipage}[b]{0.48\textwidth}
        \centering
        \begin{minipage}[b]{0.95\textwidth}
        \vspace{-2mm}
        \centering
        \hspace{2mm}
        \scriptsize
         % Table with caption at the top
        % \captionsetup{labelformat=empty}
        \captionof{table}{RPE (Trans (\%)/Rot (deg/m)) in NTU4DRadLM}
        \label{tab:result_rpe}
        \resizebox{\textwidth}{!}{  % Resizes the table to fit the width of minipage
            \begin{tabular}{l|cccc}
                \hline
                \textbf{Method} & \textbf{cp} & \textbf{nyl} & \textbf{loop2} & \textbf{loop3} \\ \hline\hline
                4DRadarSLAM & 0.106/1.107 & 0.178/1.106 & 0.745/0.751 & 0.472/0.562 \\ 
                EKF-RIO     & 0.298/1.806 & 0.423/1.864 & 1.750/1.179 & 1.307/0.596 \\
                BCV         & 0.096/1.232 & 0.142/\textbf{0.949} & 0.368/0.564 & 0.546/0.537 \\ 
                DeRO        & 0.112/1.625 & 0.318/1.610 & 2.059/0.880 & 0.373/0.524 \\ 
                Proposed    & \textbf{0.085}/\textbf{0.722} & \textbf{0.116}/1.003 & \textbf{0.338}/\textbf{0.547} & \textbf{0.365}/\textbf{0.384} \\ \hline
            \end{tabular}
        }
        \end{minipage}
        \setcounter{subfigure}{1}
        \begin{minipage}[b]{\textwidth}
        \centering
            \begin{subfigure}[b]{\textwidth}
            \vspace{3mm}
                \centering
                \includegraphics[trim=0.3cm 0.2cm 0cm 0.5cm, clip, width=1\textwidth]{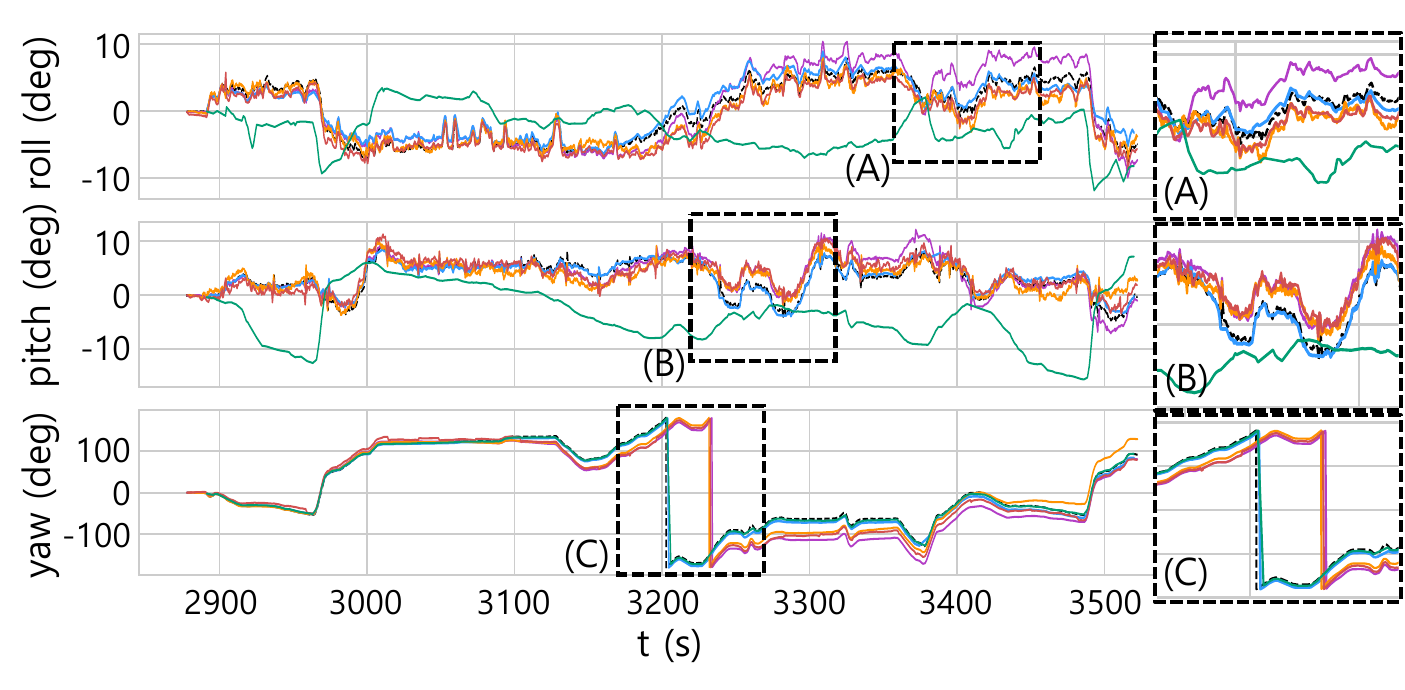}
                \vspace{-5mm}
                \caption{Estimated rotation in \texttt{loop3} (degrees)}
                \label{sugfig:ntu_loop3_rpy}
                \setcounter{figure}{4}
            \end{subfigure}
            
        \end{minipage}
    \end{minipage}
    \vspace{-1mm}
    \caption{(a) The trajectory plots of \texttt{loop2} in NTU4DRadLM. The proposed method (blue) best aligns with the ground truth (black). (b) Estimated rotation in \texttt{loop3}. In terms of heading (a zoomed view `C' in (b)), both 4DRadarSLAM and ours demonstrate superior performance; however, including the roll and pitch, ours exhibits better results (zoomed views `A' and `B' in (b)).}
    \label{fig:ntu_main}	
    \vspace{-3mm}
\end{figure*}
\begin{figure*}[!t]
    \centering
    \hspace{1mm}
     \begin{subfigure}[b]{0.2\textwidth}
        \centering
        \includegraphics[trim=1.2cm 0cm 1.5cm 0cm, clip, width=\textwidth]{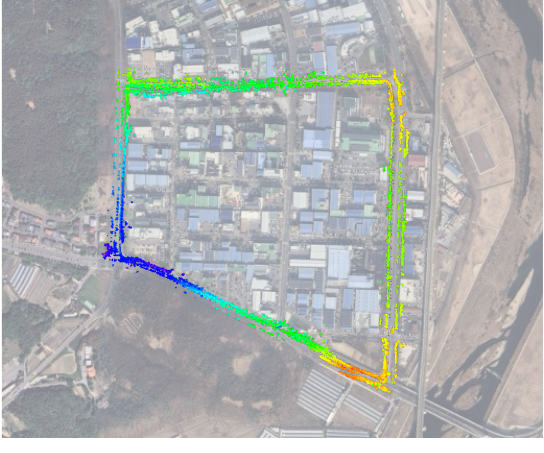}
        \vspace{-3mm}
        \caption{Map based on odometry}
        \label{subfig:msc_map}
    \end{subfigure}
    \hspace{-1mm}
    \begin{subfigure}[b]{0.43\textwidth}
        \centering
        \includegraphics[trim=0cm 0cm 4.5cm 0cm, clip, width=\textwidth]{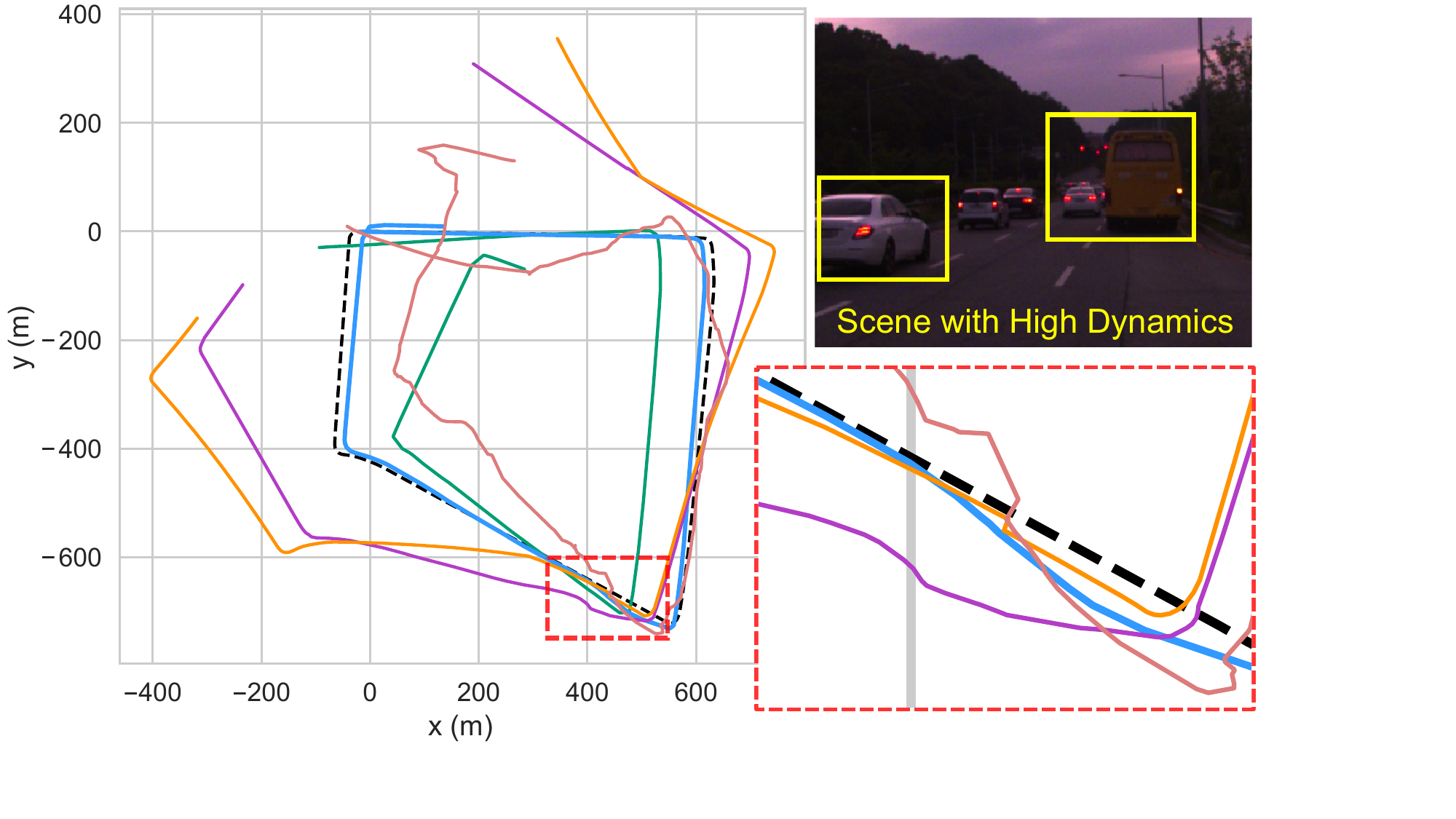}
        \vspace{-10mm}
        \caption{Estimated trajectory in \texttt{LOOP\_A0}}
        \label{subfig:msc_loopa0_traj}
    \end{subfigure}
    % \hspace{2mm}
    \begin{subfigure}[b]{0.35\textwidth}
        \centering
        \includegraphics[trim=0cm 0cm 0cm 0.2cm, clip, width=0.95\textwidth]{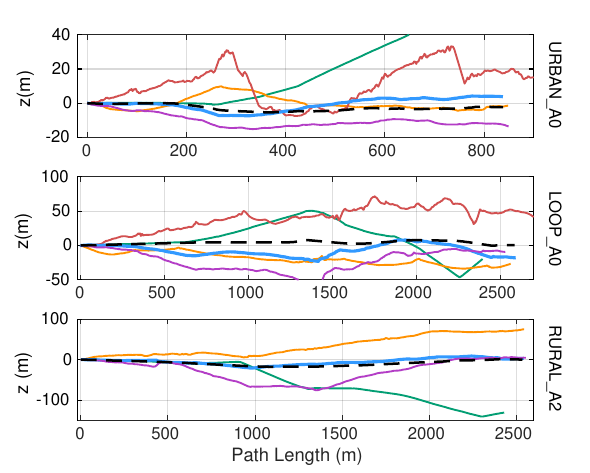}
        \vspace{-2mm}
        \caption{Elevation over traveled path length (m)}
        \label{subfig:msc_elevation}
    \end{subfigure}
    \hspace{-3mm}
    \vspace{-1mm}
    \caption{(a) Qualitative analysis of the proposed odometry in \texttt{LOOP\_A0}. Point cloud map based on our odometry shows well-alignment with the satellite image. (b) Sudden ego-velocity drift can occur due to large dynamic objects in \texttt{LOOP\_A0}. While other radar-inertial baselines fail to generate accurate trajectories, our method effectively handles the challenging scenarios. (c) Detailed analysis in elevation on \texttt{URBAN\_A0}, \texttt{LOOP\_A0}, \texttt{RURAL\_A2}, respectively. In \texttt{RURAL\_A2}, EKF-RIO is omitted for clarity.}
    \label{fig:msc}	
    \vspace{-7mm}
\end{figure*}

As detailed in \tabref{tab:result_ate}, our method represents consistent performance through the experiments.
In the cases of \texttt{nyl} and \texttt{loop2}, where noise generated by sunshades or ceilings is prevalent, 4DRadarSLAM encounters performance degradation induced by the absence of effective noise filtering.
Despite this, our method exhibits robust performance owing to ground filtering.
\texttt{loop2} involves vigorous rotational movements, discretized propagation model in EKF-RIO and BCV struggled from orientation estimation as illustrated in \figref{subfig:ntu_loop2_traj}. DeRO exhibits the worst performance, affected by both noise and discretization.
Conversely, our continuous model with \ac{GP} enables accurate estimation throughout the experiments, demonstrating that the proposed method effectively handles the aforementioned challenges.

Interestingly, 4DRadarSLAM achieves the lowest \ac{ATE} in \texttt{loop3} even without utilizing \ac{IMU}. The \texttt{loop3} sequence includes vertical structures such as walls and hills, which generate horizontal multipath.
Our filtering process, which focuses on rejecting underground outliers, is less effective for these types of noise.
Nevertheless, our method exhibits the lowest RPE$_r$, validating the effectiveness of continuous integration as shown in \tabref{tab:result_rpe} and \figref{sugfig:ntu_loop3_rpy}.

% Additionally, the integration between \ac{IMU} gyroscope and radar velocity is influenced by the precision of extrinsic calibration. NTU4DRadLM performed extrinsic calibration between radar and \ac{IMU} along with \ac{LiDAR}, RGB, and thermal cameras. As the number of sensors involved in the calibration process increases, the accumulated calibration errors become inevitable, causing discrepancies in the integration between \ac{IMU} orientation and radar velocity.

% unstructured environment DBSCAN error: clustering based weight provides wrong, 

% calibration effect : for integration, we transform the ego velocity from radar frame to imu frame, which needs the extrinsic value.(v-[w]r)
% For \ac{IMU} and radar velocity interpolation, accurate extrinsic calibration is needed.
% In NTU4DRadLM, the radar-\ac{IMU} calibration is conducted through radar - thermal - RGB - lidar - imu. 
% the calibration error might be accumulated, (especially for thermal cameras)

% loop2 loop3 contains lots of  rotation in the sequence and long, which makes whole ATE worse (4~5km)

\subsection{Evaluation on the MSC-RAD4R Dataset}
\label{subsec:msc}

In MSC-RAD4R, our method outperforms the other baselines as represented in \tabref{tab:result_ate}.
\texttt{URBAN\_A0} is characterized by numerous street trees, resulting in dominant vertical line features. These features are incompatible with point-to-plane errors in 4DRadarSLAM, causing localization failure.
In the \texttt{LOOP} sequences, which involve high vehicle speed with rapid rotational movements, 4DRadarSLAM indicates the scale misalignment.
% Although DeRO attempts to compensate for the tilt angle with \ac{IMU} accelerometer, introducing drift due to noisy \ac{IMU} measurement.
As shown in \figref{fig:msc}, large dynamic objects can produce abrupt drifts in ego-velocity estimation. Discretized propagation models that rely on constant state assumptions exacerbate these errors, as illustrated in the performance of BCV and DeRO.
Nonetheless, the proposed algorithm demonstrates consistent performance benefiting from reliance on the \ac{GP}-based continuous model, which effectively compensates for these measurement drifts and results in a smooth trajectory.
\texttt{RURAL\_A2} and \texttt{RURAL\_B2} are characterized by sharp roundabouts and high dynamics in harsh weather conditions, presenting significant challenges. Nonetheless, our algorithm maintains reliable performance.
\begin{figure*}[!t]
    \centering
    % First subfloat
    \begin{subfigure}[b]{0.51\textwidth}
        \centering
        \includegraphics[trim=0cm 10.3cm 13cm 0cm, clip, width=\textwidth]{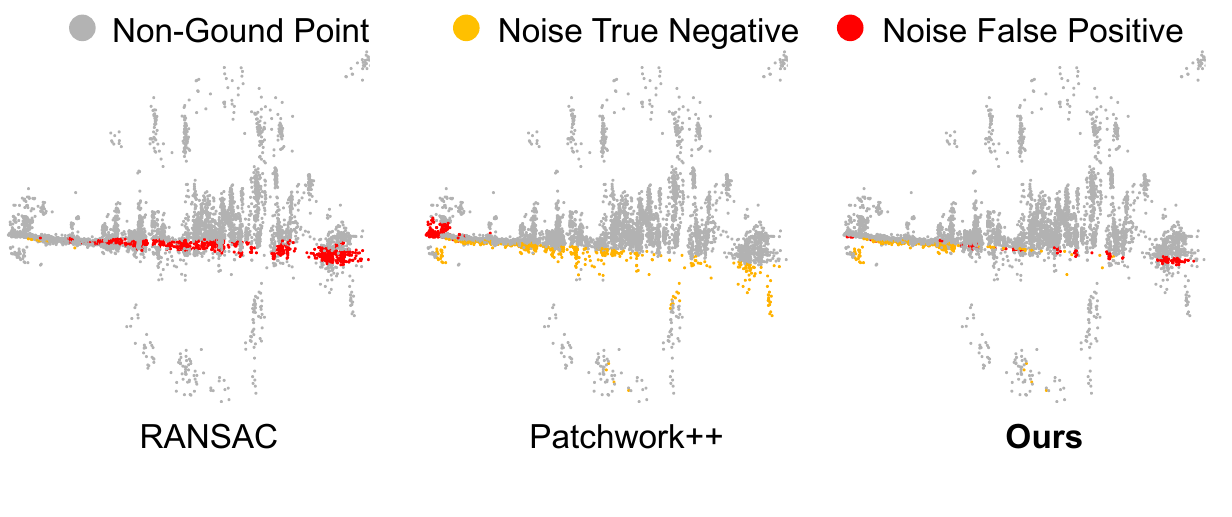}
        \vspace{-6mm}
        \caption{Noise filtering result of each method in \texttt{nyl}}
        \label{subfig:filter}
    \end{subfigure}
    \hspace{-2mm}
    % Second subfloat
    \begin{subfigure}[b]{0.48\textwidth}
        \centering
        \includegraphics[trim=0cm 0cm 0cm 0cm, clip, width=\textwidth]{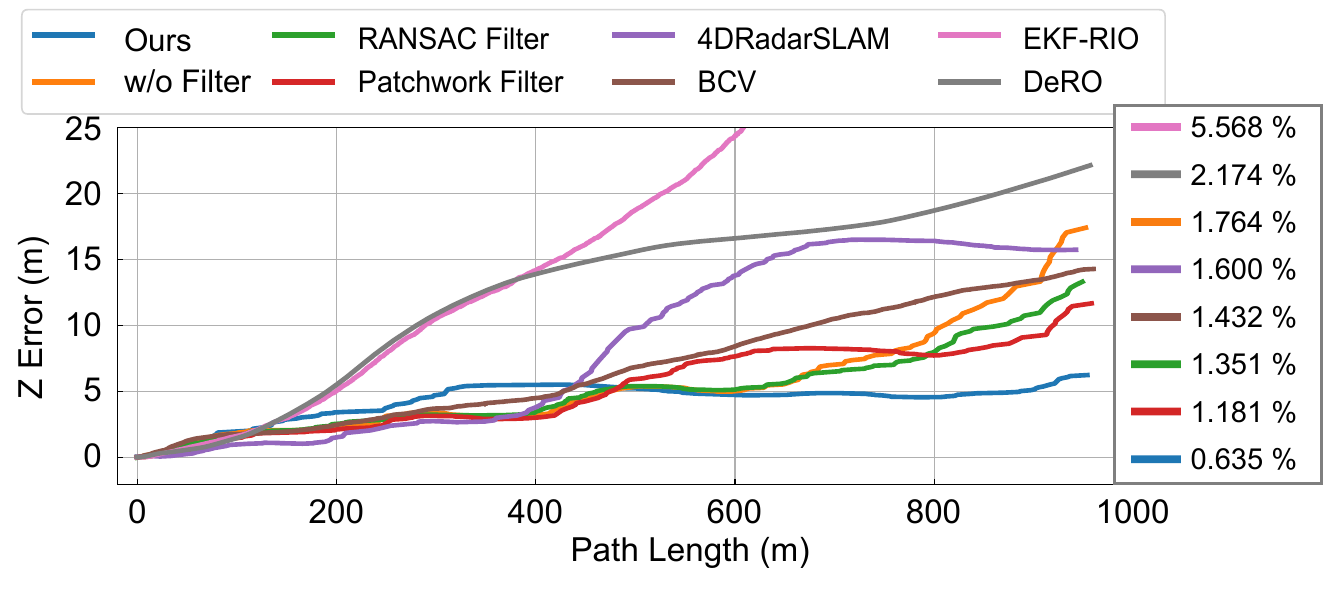}
        \vspace{-6mm}
        \caption{$z$-axis error in \texttt{nyl}}
        \label{subfig:elevation}
    \end{subfigure}
    \caption{(a) Qualitative results of each filtering method projected onto the $XZ$-plane.
    The ground truth of the noise was created by projecting the \ac{LiDAR} scan onto the radar. (b) The effect of each outlier rejection in $z$-axis odometry. Vertical errors over path length are detailed in the right box.}
    \label{fig:elevation}
    \vspace{-6mm}
\end{figure*}
Remarkably, 4DRadarSLAM achieves comparable results without leveraging \ac{IMU}, underscoring the appropriate fusion model for asynchronous sensor measurements.
Throughout the experiments, we observe that our methodology exhibits minimal elevation error.
%, which will be comprehensively explored in \secref{subsec:ablation}.

\begin{table}[t]
    \scriptsize
    \caption{Effect of Each Module in ATE}
    \vspace{-1mm}
    \label{table:ablation}
    \centering
    \resizebox{0.35\textwidth}{!}{%
    \begin{tabular}{ccccc}
    \toprule
    & \texttt{RAW} & \begin{tabular}[c]{@{}c@{}}\texttt{CONT} \end{tabular} & \begin{tabular}[c]{@{}c@{}}\texttt{FILTER} \end{tabular} & \texttt{FULL} \\
    \midrule[1pt]
    \multicolumn{1}{c|} {\texttt{cp}} & 2.337 & \textit{1.307} & 2.029 & \textbf{0.699} \\
    \midrule
    \multicolumn{1}{c|}{\texttt{nyl}} & 26.319 & 19.069 & \textit{5.888} & \textbf{5.009} \\
    \midrule
    \multicolumn{1}{c|}{\texttt{URBAN\_A0}} & 13.433 & \textit{2.163} & 12.412 & \textbf{1.757}   \\
    \midrule
    \multicolumn{1}{c|}{\texttt{LOOP\_A0}} & 101.956  & \textit{23.345} & 84.664 & \textbf{16.752}   \\
    \midrule
    \multicolumn{1}{c|}{\texttt{RURAL\_A2}} & 76.799 & \textit{20.764} & 47.966 & \textbf{9.480}   \\
    \bottomrule
    \end{tabular}}
    \vspace{-5mm}
\end{table}

\subsection{Effect of Uncertainty-Aware Ground Filtering}
\label{subsec:ablation_ground}

% To verify the effect of each module, we performed ablation studies on filtering and continuous velocity integration.
In the following analysis (\secref{subsec:ablation_ground}-\secref{subsec:ablation_velgp}), we denote our system as \texttt{FULL}, uncertainty-aware ground filtering as \texttt{FILTER}, and continuous velocity preintegration as \texttt{CONT}. We perform ablation studies to verify the effect of each module, compared to discrete integration without any noise filtering method (\texttt{RAW}).

To evaluate the effectiveness of our filtering method, we conduct a comparative analysis with the RANSAC-based naive plane fitting \cite{li20234d} and Patchwork++ \cite{lee2022patchworkpp}. The qualitative results are depicted in \figref{subfig:filter}. The RANSAC-based approach exploits height thresholding for ground point segmentation; however, this produces false positives in areas with varying terrain elevation, such as slopes, resulting in the erroneous classification of significant static features as noise. Patchwork++ fails to account for spatial uncertainties associated with distant points, yielding unmitigated noise despite utilizing zone-based ground segmentation. In contrast, our method accurately filters out the noises.

\Cref{subfig:elevation} indicates that our approach outperforms other methods over elevation accuracy.
\tabref{table:ablation} presents the quantitative improvements in ATE attributed to our filtering method.
As expected, \texttt{RAW} shows diminished performance compared to \texttt{FILTER} due to the absence of noise filtering, demonstrating the robustness of our filtering approach in varying conditions.
These improvements are particularly significant in the \texttt{nyl} dataset, where the majority of noise consists of underground noise caused by sunshades. This denotes that our method is highly effective in mitigating this predominant type of noise.
In conclusion, our uncertainty-aware ground model effectively handles noises and enhances the odometry accuracy.

\subsection{Effect of Continuous Velocity PreIntegration}
\label{subsec:ablation_velgp}

In \tabref{table:ablation},
\texttt{FULL} demonstrates superior performance over \texttt{RAW} and \texttt{FILTER}, both numerically integrate the \ac{IMU} angular velocity and the radar velocity \cite{kubelka2024do}. 
The discretized propagation model to address the temporal discrepancies exhibits limitations due to measurement drift and constant state assumptions, resulting in significant degradation on \texttt{LOOP\_A0} and \texttt{RURAL\_A2} with high vehicle speeds and sharp turns.
Conversely, \texttt{FULL} effectively accommodates vigorously changing orientations and velocities, attributed to the proposed velocity preintegration, which enables direct motion estimation through asynchronous \ac{IMU} angular velocities and radar velocities.
Moreover, the smoothness nature of \ac{GP} makes our preintegration process more robust to unavoidable measurement noise.

%TABLE
\begin{table}[t]
    \scriptsize
    \caption{Computation Time of Each Module (ms)}
    \vspace{-1mm}
    \label{table:time}
    \centering
    \resizebox{0.45\textwidth}{!}{%
    \begin{tabular}{cccccc}
    \toprule
    & Ego-vel & \begin{tabular}[c]{@{}c@{}}Ground \end{tabular} & Scan Matching & GP Integration & Optimization \\
    \midrule[1pt]
    \multicolumn{1}{c|}{\multirow{1}{*}{\texttt{loop2}}} & 0.31 & 68.38 & 2.87 & 615.43 & 370.45 \\
    \midrule
    \multicolumn{1}{c|}{\multirow{1}{*}{\texttt{LOOP\_A0}}} & 0.99 & 52.67 & 2.05 & 789.77 & 436.77 \\
    \bottomrule
    \end{tabular}}
    \vspace{-5mm}
\end{table}
%TABLE

\subsection{Computational Cost}
\label{subsec:timecost}

The time consumption analysis results in \texttt{loop2} and \texttt{LOOP\_A0}, which represent the longest paths in the dataset, are presented in \tabref{table:time}. The experiments were conducted on Intel i7 CPU@2.50 {\GHz} and 64 {\GB} RAM.
Uncertainty-aware ground filtering and scan matching can be executed in real-time. While \ac{GP} integration and optimization are computationally expensive, they are performed in parallel in the back-end, ensuring that our system maintains real-time capability. Moreover, additional performance enhancements can be pursued during continuous preintegration.

\section{Conclusion}
\label{sec:conclusion}

In this paper, we present a tightly-coupled 4D-radar-inertial odometry framework. By employing uncertainty-aware ground filtering, we effectively eliminate the inherent noises from radar. Continuous velocity preintegration via \ac{GP} effectively mitigates temporal discrepancies between radar and \ac{IMU}, and enables robust motion estimation directly from the asynchronous measurements without any assumptions.
% Additionally, spatial information is incorporated through cluster-based weighted scan matching.
We have demonstrated that our method outperforms existing 4D radar-inertial odometry algorithms in terms of precision and resilience across various scenarios through the public datasets.
Notably, the proposed algorithm achieves substantially enhanced vertical odometry accuracy.

Despite significant advancements, minor challenges remain. Since our filtering process primarily targets underground noise, horizontal multipath remains a concern. For future work, we aim to incorporate vertical plane modeling to enhance the robustness of the odometry and mapping system. Also, we plan to improve our system by developing a comprehensive bias model for the IMU gyroscope or incorporating the accelerometer measurements.
% acceleration integration for system enhancement?

\newpage

% \newpage
% \newpage

%\section*{ACKNOWLEDGMENT}
\balance
\small
\bibliographystyle{IEEEtranN} %citeauthor
\bibliography{string-short,references}

\end{document}